# Face Photo Sketch Synthesis via Larger Patch and Multiresolution Spline

Xu Yang

xuyangaca@gmail.com

**Abstract:** Face photo sketch synthesis has got some researchers' attention in recent years because of its potential applications in digital entertainment and law enforcement. Some patches based methods have been proposed to solve this problem. These methods usually focus more on how to get a sketch patch for a given photo patch than how to blend these generated patches. However, without appropriately blending method, some jagged parts and mottled points will appear in the entire face sketch. In order to get a smoother sketch, we propose a new method to reduce such jagged parts and mottled points. In our system, we resort to an existed method, which is Markov Random Fields (MRF), to train a crude face sketch firstly. Then this crude sketch face sketch will be divided into some larger patches again and retrained by Non-Negative Matrix Factorization (NMF). At last, we use Multiresolution Spline and a blend trick named full-coverage trick to blend these retrained patches. The experiment results show that compared with some previous method, we can get a smoother face sketch.

**Key words:** face synthesis; probabilistic graph model; nonnegative matrix factorization; multiresolution spline.

## 1 Introduction

Face photo sketch synthesis has attracted more and more attention in recent years. The development of this technique facilitates many applications such as the law enforcement and digital entertainment.Researchers have proposed various methods for photo-sketch synthesis in recent years. [7] used Hough transform to generate facial sketch of line drawings. A series of filters were used in [8] and [9] to transform a face photo to a face sketch. These methods didn't depend on any training set and they generated a sketch directly. So the results of these methods didn't contain any distinctive characteristics of sketch painting.

[10] and [11] used Principal Component Analysis (PCA) to synthesize and retrieve face sketch. The algorithms of these papers can learn some distinctive texture of sketch painting with the resort of the training set of photo-sketch pairs. However, these algorithms were applied to the whole face image except the hair region, which can also provide indispensable information in some applications. Moreover, PCA works in the whole face region will cause the generated sketches to be oversmoothed.

Patches-based method was proposed in [1], [2], and [12]-[21]. Patches-based method usually includes three steps. Firstly, a face photo is divided into some overlapping small patches. Secondly, for each small photo patch, a corresponding sketch patch will be generated. At last, all these small sketch patches will be blended into an entire face sketch. There are two justifications for dividing a whole photo into small patches. The first reason is to reduce the complexity of converting a photo to a sketch. The second one is that for a small patch, more details can be sought compared with a whole picture.

[2], [12]-[14] exploited the technique of sparse representation to generate each sketch patch from the corresponding photo sketch. A Markov Random Field model was built in [1] to learn sketch patch, [15] developed a new model called Markov Weight Field from [1] to produce sketch patch. In addition, some other regression methods were used in [16]-[17]. The computer simulation results of these papers demonstrated their effectiveness in generating sketch patches from the given photo patches.

However, the focuses of these algorithms were primary on learning perfect patches. They paid little attention to the strategy of blending, which is also important in generating a smooth face sketch. In [2], [12]-[14], [15], [16] and [21], the pixel values in overlapping region were averaged, which will maculate the generated face sketch with blurring effect. Minimum error boundary cut strategy, which was proposed in [5], was used in [1] which can eliminate blurring effect but instead will cause jagged artifacts in synthesized sketch. Another limitation of these previously algorithm is that by dividing a whole face photo into small patches, some global or long-range information will lose [1]. The sketch drew by a human artist will contain some long-range shadow textures to show 3D effect. And these textures will be loss if the divided patches are too small.

The photo-sketch synthesis method proposed by this paper is aim at ameliorating these two disadvantages. 1. Obtaining local details and long-range information simultaneously. 2. Reducing the blurring effect or jagged artifacts when combining disjunctive patches. More specifically, MRF with smaller patches and NMF retaining with larger patches can make results preserve local details and long-range information simultaneously. Multiresolution Spine and full-coverage blend trick will be used to reduce the blurring effect and jagged artifacts.

Our framework includes three steps. Firstly, MRF, which is proposed in [1], is used to generate a crude face sketch. Secondly, NMF is used to retrain this crude face sketch with larger patches (section II). At last, these retrained larger patches will be blended into an entire face sketch with Multiresolution Spine and full-coverage blend trick ((section III).

## 2 Crude Sketch Generating and NMF retraining

We generate a crude sketch using the method proposed in [1]. In this method, the photo is divided into some small overlapping patches. The size of the patch is chosen to be $10 \times 10$. For each photo patch, there are two constraints for the corresponding sketch patch. The first one is that the target sketch patch should be similar to the original photo patch. The second one is that the neighboring sketch patches should be compatible. This compatible constraint requires that the differences of pixel values in an overlapping region within two neighboring sketch patches should be as small as possible. Then a MRF model is built and Belief Propagation algorithm can be used to solve this model [4]. Readers can get more detail about MRF model in [1].

After getting a crude sketch from the MRF model, NMF will be used to retrain this crude face sketch [6]. In our experiments, we divide the crude face sketch into $N$ overlapping patches and denote these patches as $\{y\}_{i=1}^{N}$. Note that the size of patch $y_i$ is $20 \times 20$ and the value of overlap is 10. The reason for this will be discussed later when the full-coverage blending trick is proposed.

Then $N$ NMF dictionaries will be trained in our model, each one for different spatial location to gain enough detail features. The same division strategy is used in the sketches drew by artist.

Assume there are $M$ sketches in training set, so for each spatial location there are total $M$ patches with dimension $d$ (in this special case $d = 20 \times 20 = 400$). Patches can be treated as column vectors of the data matrix. By NMF this data matrix will be decomposed to the product of two non-negative matrixes. We denote the $d \times M$ matrix (data matrix) as $V_i$, the subscript $i$ index the spatial location of patch $i$ in the whole sketch. The data matrix $V_i$ is then approximately factorized into a $d \times r$ matrix $W_i$ and an $r \times M$ matrix $H_i$. $W_i$ and $H_i$ should satisfy non-negative constraint meantime. The NMF problem can be formulated as follow.

$$\min_{W_i, H_i} \|V_i - W_i H_i\|$$
$$s.t.\ V_i \geq 0, H_i \geq 0 \quad (1)$$

The update rule of (1) was given in [6] as follows:

$$H_{ij} \leftarrow H_{ij} \frac{(W^T V)_{ij}}{(W^T W H)_{ij}}, W_{jk} \leftarrow W_{jk} \frac{(V H^T)_{jk}}{(W H H^T)_{jk}} \quad (2)$$

Fig. 1 shows one case of this process, in which the patch of the left eye of face is used to illustrate NMF training.

After training $W_i$ and given the crude sketch generated in the MRF model.

The sketch patches with more long-range information can be trained though the following optimization:

$$\min_{\alpha_i} \|y_i - W_i \alpha_i\| \quad (3)$$

The target sketch patch is $y_i^t = W_i \alpha_i$. Minimizing (3) targets at making retained sketch to be similar to the crude one. And taking $y_i^t$ as a sum of column vectors in $W_i$ is to guarantee the retained sketch to preserve special information of sketches drew by artist.

## 3 Multiresolution Spline And Full-Coverage Trick

After NMF retaining, scattered patches need to be combined into a whole sketch face. As mentioned above (in introduction), if the improper blending method is chosen, blurring effect, visible edge or jagged artifacts will appear. However, multiresolution spline, which was proposed in [3], can blend few images into one with fewer visible edge compared with averaging pixel value or minimum error cut. Multiresolution Spline modifies image gray levels in a transition zone whose width is different for diverse spatial frequency component images. Thus the mutation of overlapping region of two images will be relieved, which consequently reduces the effect of visible edge.

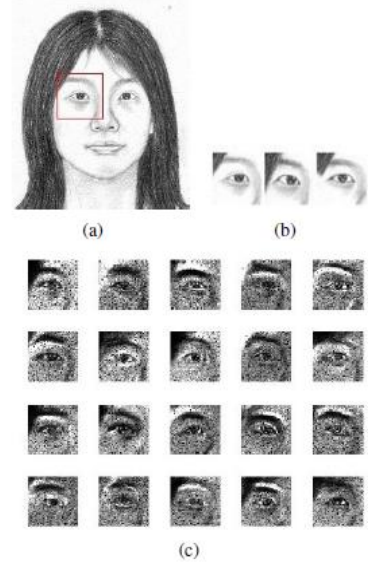

Fig. 1. Illusion of NMF for patches of left eye. (a) The red box shows the location of the extractive patch in each sketch drew by artist. Patches of left eye are used to illustrate this process here. (b) Three patches which are extracted from three different sketches. In this illusion, there are totally 100 patches are extracted from 100 sketches in the training set. (c) This picture shows 20 atoms in dictionary $W$ learned in NMF, where the parameter $r$ is set as 20.

It is clear that if two images are similar, the blended image will be smoother. For this reason, we propose a blend trick named full-coverage trick to make two patches as similar as possible. Here we will use a simple example to show this trick. Fig. 2 shows this trick. Suppose the size of face sketch generated in the MRF model is $40 \times 40$. Then this sketch is divided into the overlapping patches (the size is $20 \times 20$), with the value of overlap to be 10, so totally $\frac{(40-10) \times (40-10)}{(20-10) \times (20-10)} = 9$ patches can be got. For each $20 \times 20$ patch, NMF is used to learn an enhancement patch, and these patches are denoted by $\{y_i^t\}_{i=1}^9$.

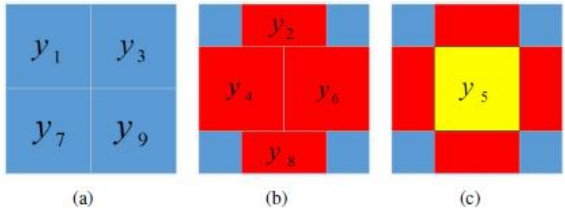

Fig. 2. (a) The first step in the blend trick, $y_1^t, y_3^t, y_7^t, y_9^t$ are stitched. (b) The second step in the blend trick, $y_2^t, y_8^t, y_4^t, y_6^t$ are blended into the image stitched in the first step. (c) The last step in the blend trick, in which $y_5^t$ is blended into the image got in the second step.

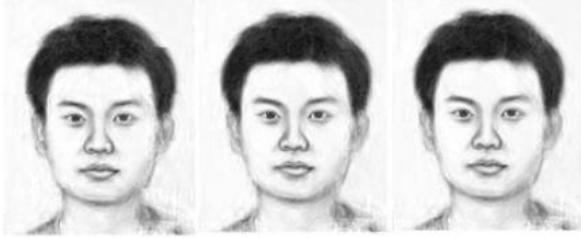

Fig. 3. One example of results of each step in full coverage blend trick. In order to make the edge to be more visible, the size of patch is set as $50 \times 50$. The left one shows that after the first step of our blend strategy, some visible edges appear. The middle one is got after second step, which is almost equivalent to the ultimate result, expect some fleck points appear. And the right one is got from the entire process, the visible edges and fleck points are eliminated.

Firstly, the patches without any overlapping region will be stitched in the full-coverage blend trick.

In this example $y_1^t, y_3^t, y_7^t, y_9^t$ will be set in the location (0,0), (0,20), (20,0), (20,20) respectively ((the coordinates here are the ones of upleft corners of patches, and the origin is the upleft corner)).

In Fig.2 these patches are indexed by blue. Visible edges will appear between the neighboring patches after the first step. And these edge will be eliminated after the second step, in which we blend $y_2^t, y_8^t, y_4^t, y_6^t$ in the location (0,10), (20,10), (10,0), (10,20) with the image got from the first step. These patches are indexed by red. At last, patch $y_5^t$ will be blended into (10,10) to eliminate the fleck point appeared in (20,20) and this patch is indexed by yellow. Note that the blend method used here is the Multiresolution Spine. One case of experiment results of each blend step are demonstrated in Fig. 3.

Note that the size of patch is chosen to be larger,

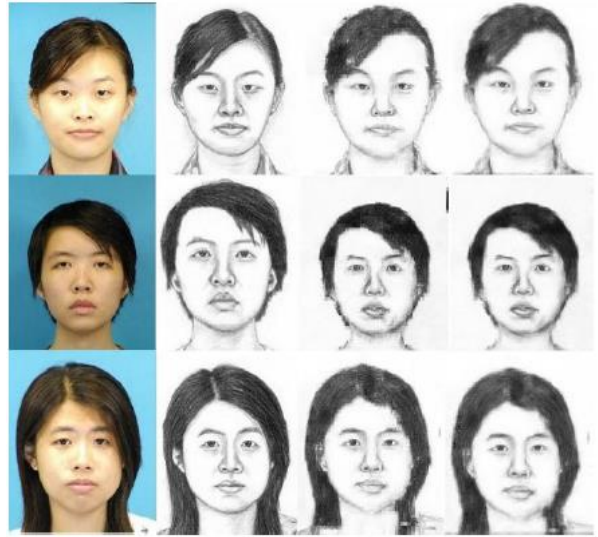

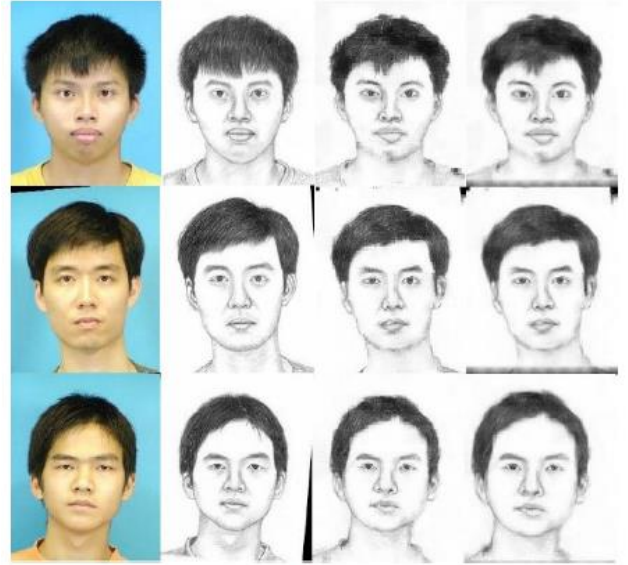

Fig. 4. Some examples of generated sketches. From left to right are: (1) face photos; (2) sketches drew by artists; (3) the crude results generated in the first step and (4) the ultimate results of the whole process.

which is $20 \times 20$, can be justified from three different perspectives. Firstly, from the perspective of NMF retraining. In the NMF model, if the patch is so large, then the last generated sketch will be over-averaged as Fig. 3 shows (in which the size is $50 \times 50$). So the size should not be so large. Secondly, from the perspective of Multiresolution Spline method. As discussed in [3], the width of the transition region is important. And if the patch is so small, the width for different spatial frequency component will not change so much. Then the Multiresolution Spline will degenerate to single-resolution spline. Then the generated sketch will thus have some jagged parts, as Fig. 5(b) shows. So the size of patch should not be so small. Thirdly, from the perspective of full coverage trick. Because the size of face photo is $250\times 200$, then if the size of patch is $20\times 20$ and overlap is 10, we can get integral number of patches, which is 50.

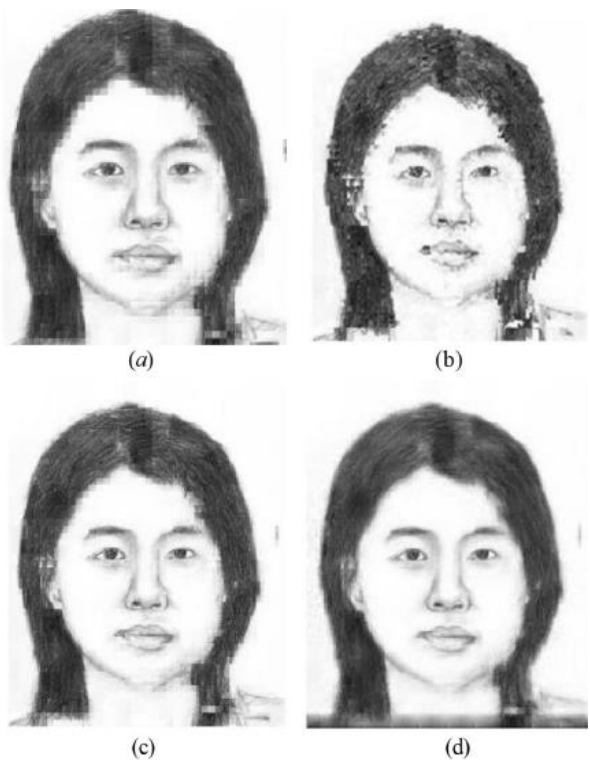

Fig. 5. Comparisons between different blend strategy.

## 5 Result and analysis

Fig.4 demonstrates some synthesized face sketches in different stage of the proposed framework. As the results show, generated sketches eliminate some jagged parts and make the sketch be smoother.

Fig. 5 compares four different blend strategy of some previously algorithm used. Fig. 5(a) shows the sketch which got by averaging the overlapping pixels. As the figure shows, apparent jagged parts appear in the margin region between hair and face. Minimum error boundary cut is used in (b). The sketch is full of mottled points in this picture. Multiresoultion Spline and full-coverage blend trick is used in (c) and (d). The only difference between (c) and (d) is the size of patch: the size of patch is $10\times 10$ and $20\times 20$ in (c) and (d) respectively. (c) shows less jagged parts and mottled points compared with (a) and (b). When the size of patch is larger, as shown in (d), the generated face sketch is smoother than (a), (b) and (c). In addition, the jagged parts and mottled points are almost eliminated in (d).

## 6 Conclusion

In this paper, we proposed a noval framework to generate a face sketch from a given face photo. Compared with some existed methods, we focus more on how to get a smoother entire face sketch from the crude one. Larger size patches and Multiresolution Spline were used to enhance a crude sketch generated from one existed method. Our experiment results showed that after enhancing, the ultimate was better than the crude ones.